\documentclass[11pt,a4paper]{article}
\usepackage[hyperref]{acl2020}
\usepackage{times}
\usepackage{latexsym}

\usepackage{linguex} 
\usepackage{comment} 
\usepackage{graphicx}
\usepackage{gb4e}

\usepackage[11pt]{moresize}
\noautomath
\usepackage{microtype}
\usepackage{xcolor}
\usepackage{booktabs}

\newcommand{\brackets}[1]{$\langle\rangle$}
\newcommand{\argn}[1]{\underline{#1}$_{ARG}$}
\newcommand{\arga}[1]{\underline{#1}$_{ARG_1}$}
\newcommand{\argb}[1]{\underline{#1}$_{ARG_2}$}
\newcommand{\argas}[1]{\underline{#1}$_{A_1}$}
\newcommand{\argbs}[1]{\underline{#1}$_{A_2}$}

\newcommand{\cameraReadyFootnote}[1]{}

\aclfinalcopy 

\title{Neural Extractive Search} 

\author{Shauli Ravfogel\textsuperscript{1,2} \, Hillel Taub-Tabib\textsuperscript{2} \, Yoav Goldberg\textsuperscript{1,2}\\
\textsuperscript{1}Computer Science Department, Bar Ilan University \\
\textsuperscript{2}Allen Institute for Artificial Intelligence \\
  {\tt  \{shauli.ravfogel, yoav.goldberg\}@gmail.com} \\ \tt {hillelt@allenai.org} 
  }

\date{}

\begin{document}
\maketitle
\begin{abstract}
Domain experts often need to extract structured information from large corpora.  We advocate for a search paradigm called ``extractive search'', in which a search query is enriched with capture-slots, to allow for such rapid extraction. Such an extractive search system can be built around syntactic structures, resulting in high-precision, low-recall results. We show how the recall can be improved using neural retrieval and alignment. The goals of this paper are to concisely introduce the extractive-search paradigm; and to demonstrate a prototype neural retrieval system for extractive search and its benefits and potential.
Our prototype is available at \url{https://spike.neural-sim.apps.allenai.org/} and a video demonstration is available at \url{https://vimeo.com/559586687}.
\end{abstract}

\section{Introduction}
In this paper we demonstrate how to extend a search paradigm we call ``extractive search'' with neural similarity techniques.

The increasing availability of large datasets calls for search tools which support different types of information needs. Search engines like Google Search or Microsoft Bing are optimized for surfacing documents addressing information needs that can be satisfied by reviewing a handful of top results. Academic search engines (Semantic Scholar, Google Scholar, Pubmed Search, etc) address also information needs targeting more than a handful of documents, yet still require the user to read through the returned documents. 

However, some information needs require 
\emph{extracting} and \emph{aggregating} sub-sentence information (words, phrases, or entities) from multiple documents (e.g. a list of all the risk factors for a specific disease and their number of mentions, or a comprehensive table of startups and CEOs). These typically fall outside the scope of search engines and instead are classified as Information Extraction (IE), entailing a research project and a dedicated team per use-case, putting them well beyond the abilities of the 
typical information seeker.

\begin{figure*}[t!]
\centering
\includegraphics[width=0.99\textwidth]{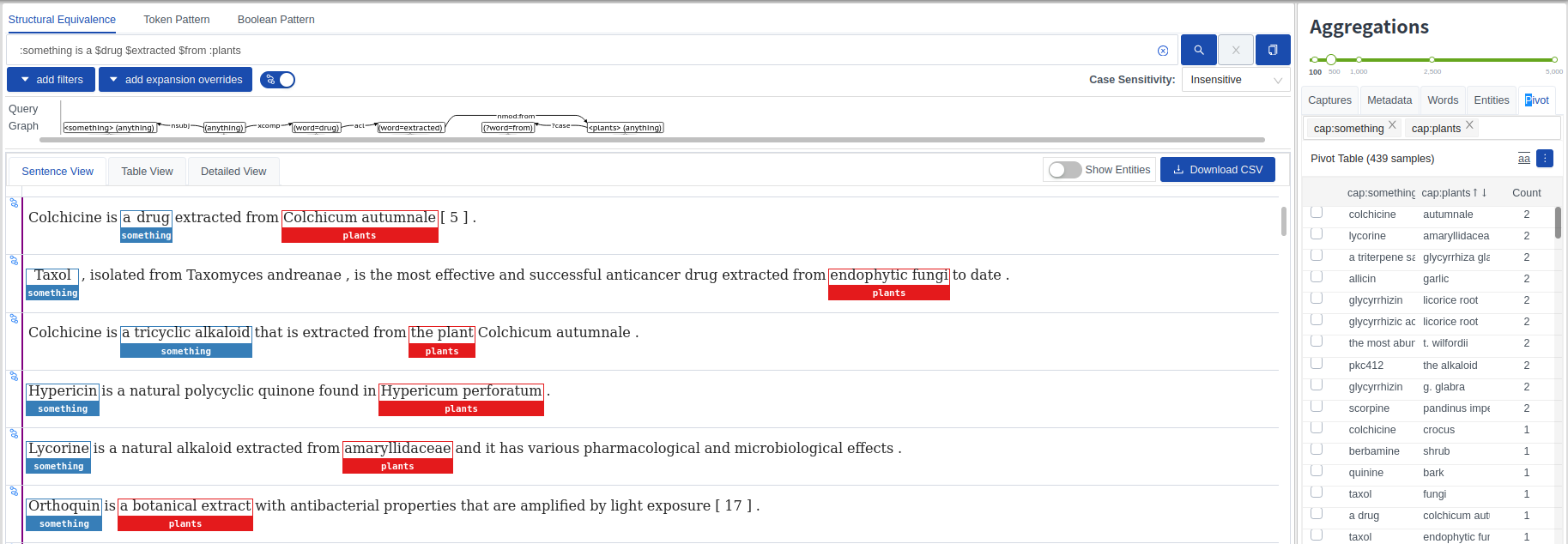}
\caption{Results of neural extractive search. The neural results are based on the syntactic query:\\ 
{\centering \emph{\arga{Something} is a \textbf{drug extracted from} \argb{plants}}} (underlines denote named capture slots, and bold text denotes an exact lexical match). The results show linguistic and lexical diversity w.r.t to the initial query, and highlight also spans corresponding to $ARG_1$ and $ARG_2$ (in light blue and yellow). The right box contains an aggregate view of the captured spans over many results.}
\label{fig:screenshot}
\end{figure*}

In contrast, we advocate for a complementary search paradigm: \emph{extractive search}, which combines document selection with information extraction. 
The query is extended with \emph{capture slots}: these are search terms that act as variables, whose values should be \emph{extracted} (``captured'').\footnote{Capture-slots can be thought of as being analogous to captures in regular-expressions.}
The user is then presented with the matched documents, each annotated with the corresponding captured spans, as well as aggregate information over the captured spans (e.g., a count-ranked list of the values that were captured in the different slots). 
The extractive search paradigm is currently implemented in our SPIKE system.\footnote{\url{https://allenai.github.io/spike/}} Aspects of its earlier versions are presented in \citet{spike, spike2}. One way of specifying which slots to capture is by their roles with respect to some predicate, semantic-frame, or a sentence. In particular, the SPIKE system features syntax-based \emph{symbolic extractive search}---described further in section \ref{sec:symbolic-extractive-search}---where the capture slots correspond to specific positions in a syntactic-configuration (i.e., ``capture the subject of the predicate \emph{founded} in the first capture slot, and the object of the predicate in the second capture slot''). These are specified using a ``by-example'' syntax \cite{spike}, in which the user marks the predicate and capture slots on a provided example sentence, and the syntactic configuration is inferred.

While such parse-based matching can be very effective, it also suffers from the known limitations of symbolic systems:
 it excels in precision and control, but often lacks in recall. In this work, we demonstrate how the symbolic system can be combined with the flexibility of \emph{neural} semantic similarity as induced by large pre-trained language models. Figure \ref{fig:screenshot} presents an overview of the system, containing a query with capture slots, the derived syntactic query, the returned (neural) results with marked spans, and an aggregate summary of the extracted pairs.

By allowing fuzzy matches based on neural similarity search, we substantially improve recall, at the expense of some of the precision and control.

The incorporation of neural similarity search requires two stages: retrieval of relevant sentences, and locating the roles corresponding to the capture-spans on each sentence. We use standard dense passage retrieval methods for the first part (section \ref{sec:nes}), and present a neural alignment model for the second part (section \ref{sec:alignment}).
The alignment model is generic: it is designed to be pre-trained once, and then applied to every query in real time. This allows to provide an \emph{interactive} search system which returns an initial response in near real-time, and continues to stream additional responses.

The purpose of this paper then is twofold: first, it serves as a concise introduction of the extractive-search paradigm. Second, and more importantly, it demonstrates an incorporation of neural similarity techniques into this paradigm. 

\section{Symbolic Extractive Search}
\label{sec:symbolic-extractive-search}
We introduce the extractive search paradigm through usage examples.
\paragraph{Boolean Extractive Search.}
Consider a researcher who would like to compile a list of treatments to Bacteremia (bloodstream infection). 
Searching Google for ``Bacteremia treatment'' might lead to a Healthline article discussing a handful of treatments.\footnote{https://www.healthline.com/health/bacteremia}, which is not a great outcome. A similar query in PubMed Search leads to over 30,000 matching papers, not all are relevant and each including only nuggets of relevant information. Compare this with the extractive boolean query:
\begin{center}
\emph{Bacteremia treatment :entity=CHEMICAL}
\end{center} in SPIKE-PubMed \cite{spike2}, a search system over PubMed abstracts. ``entity=CHEMICAL'' indicates that we are interested in spans that correspond to chemicals, and the preceding colon (``:'') designate this term as a \emph{capture}. The query retrieves 1822 sentences which include the word \emph{Bactermia}, the word \emph{treatment} (added to establish a therapeutic context) and a CHEMICAL entity. The user interface also displays the ranked list of 406 different chemicals captured by the query variable. The researcher can click each one to inspect evidence for its association with Bacteremia, quickly arriving at a clean list of the common therapeutic compounds. 

\paragraph{Syntactic Extractive Search (``by example'').}
In the previous example, the capture slot was based on pre-annotated span level information (``named entities"). While very effective, it requires the entity type of interest to be pre-annotated, which will likely not be the case for most entity types. Additionally, the search is rather loose: it identifies any \emph{chemical} in the same sentence of the terms ``Bactermia'' and ``treatment'', but without establishing a semantic connection between them. What can we do when the entity type is not pre-annotated, or when we want to be more specific in our extraction target? One option is to define the capture slots using their syntactic sentential context. For example, consider a researcher interested in risk factors of stroke.
An example of this relation is given in the syntactic configuration:
\includegraphics[width=0.48\textwidth]{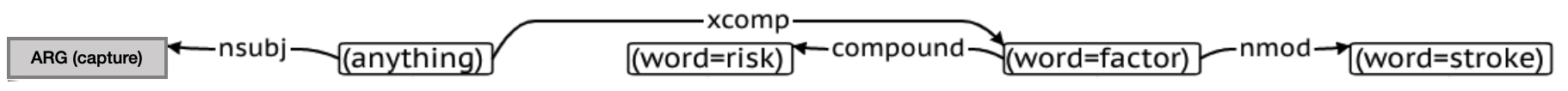}

We can search for sentences that match this pattern,\footnote{Potentially with additional restrictions such as the occurrence of other words, phrases or patterns in the document} and extract the information which aligns with the capture node.\footnote{This mode of operation is facilitated also by, e.g., the open-source toolkit Odinson \cite{odinson}, and similar workflows are discussed by \citet{propminer,extreme}.} However, such syntactic patterns require expertise to specify and are challenging to master. To counter this, \citet{spike} introduced to SPIKE the notion of query by example: the user enters a sentence which demonstrates the configuration: ``\emph{something is a risk factor of stroke}'', marks which words are essential and should match exactly (\emph{risk, factor, stroke}), and which correspond to capture slots (\emph{something}), resulting in the query:\footnote{In this paper, we avoid the exact SPIKE syntax, and use underlines to indicate named capture slots, and bolded words to indicate exact matches. The corresponding SPIKE query would be ``$\langle\rangle$ARG:something is a \$risk \$factor for \$stroke''.}
\begin{center}
 \emph{\argn{something} is a \textbf{risk} \textbf{factor} for \textbf{stroke}}
\end{center}
The system then derives the corresponding syntactic query (see \cite{spike} for the details), returning results like: 
``\emph{These cases illustrate that \underline{PXE} is a rare but significant \textbf{risk factor} for small vessel disease and \textbf{stroke} in patients of all age groups.}'', with the top aggregate terms being \emph{Hypertension, Artial fibrillation, AF, Diabetes, Obesity} while less frequent terms include \emph{VZV reactivation} and \emph{palmitic acid}. By modifying the query such that stroke is also marked as a capture slot:
\begin{center}
\emph{\arga{something} is a \textbf{risk} \textbf{factor} for \argb{stroke}}
\end{center}
one could easily obtain a table of risk factors for various conditions.

\section{Neural Extractive Search}
\label{sec:nes}
The syntactic search by example lowers the barriers for IE: it easy to specify, accurate and effective. However, it is also limited in its recall: it considers only a specific configuration (both in terms of syntax and lexical items), and will not allow for alternations unless these are explicitly expressed by the user. Neural models, and in particular large pre-trained language models \cite{bert,scibert}, excel at this kind of fuzzier, less-rigid similarity matching. We show how to incorporate them in the extractive search paradigm. This requires two stages: first, we need to match relevant sentences for a given query. Second, we need to identify the relevant capture spans in the returned sentences.
Crucially, this needs to be done in a reasonable time: we do not have the luxury of re-training a model for each query, nor can we afford to run a large neural model on the entire corpus for every query. We \emph{can} afford to run a pre-trained model on the query sentence(s), as well as over each of the sentences in the result set (similar to neural-reranking retrieval models \cite{neural_ranking}). We operate under these constraints.


The final system enables the user to search for specified information with minimal technical expertise. 
We demonstrate this approach on the CORD corpus \citep{DBLP:journals/corr/abs-2004-10706}, a collection of research papers concerning the COVID-19 pandemic. 

\subsection{`By-example'' neural queries} 
The core of the system is a ``by-example'' query, where the user enters a simple sentence expressing the relation of interest, and marks the desired capture roles on the sentence. To facilitate effective neural search based on the short example, we perform symbolic (syntactic) search that retrieves many real-world sentences following the syntactic pattern. The result is a list of sentences that all satisfy the same relation, which are then combined and used as query to the neural retrieval system. At neural alignment model is then used to align the role marking on the syntactically-retrieved sentences, to corresponding roles on the neurally-retrieved sentences.

\subsection{Pipeline}

\begin{figure*}
    \centering
    \includegraphics[width=1.8\columnwidth]{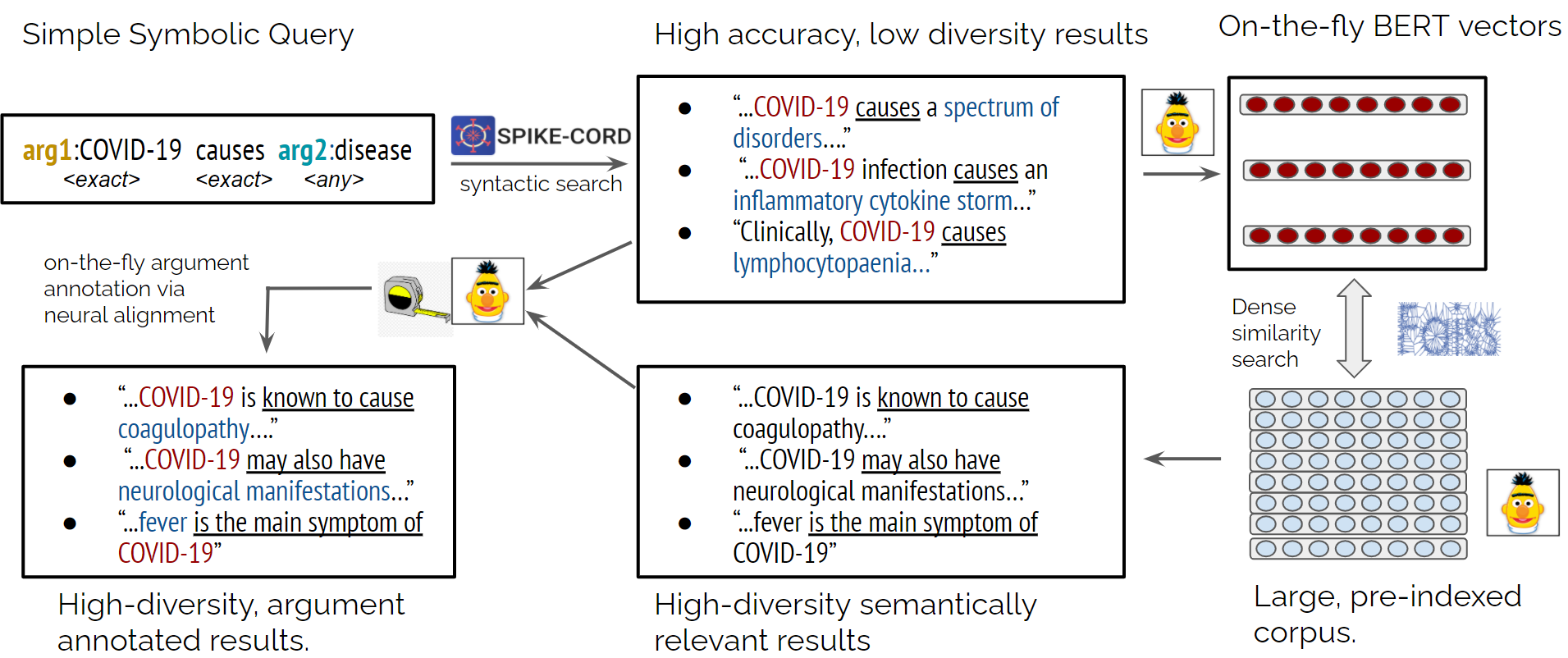}
    \caption{The proposed pipeline, presented from top left clockwise. Top: A simple symbolic query with two argument marks is provided. The query is executed, yielding accurate results that suffer from low recall. Those are encoded by BERT and used for k-NN query over a large set of pre-indexed vectors. Bottom: The k-NN neural similarity search results in a diverse set of relevant sentences.  An alignment model then predicts and annotates argument spans over the retrieved sentences, based on the symbolic query results.}
    \label{fig:pipeline}
\end{figure*}

Our system pipeline is summarized in Figure \ref{fig:pipeline}. It includes the following steps.

\paragraph{Index Construction.} Given a corpus $D=\{s_1, s_2, \dots, s_n\}$ of $n$ sentences, we calculate a vector representation $M(s_i)$ for each sentence using a neural model $M$, and index them to allow efficient search.\footnote{Concretely, we encode each sentence in the CORD-19 corpus using the pre-trained SciBERT model \cite{scibert}, a BERT-based model \cite{bert}  trained on scientific text. We do not finetune the pre-trained model. We represent each sentence by the \textit{[CLS]} representation on layer-12 of the model, and perform PCA to retain 99\% of the variance, resulting in 601-dimensional vectors.  To allow efficient search over the approximately 14M resulting dense vectors, we index them with FAISS \cite{DBLP:journals/corr/JohnsonDJ17}.}

\paragraph{Symbolic Query Encoding.}


We use the syntactic-query capabilities of the SPIKE system to retrieve examples of natural sentences that convey the meaning the user aims to capture: we collect the first 75 results of a simple ``by-example'' syntactic query as described in \S\ref{sec:symbolic-extractive-search}---which often contain lexically-diverse, but semantically coherent, sentences---and average their BERT representations in order to get a single dense query vector $\vec{h^q}$. 
The averaging helps focus the model on the desired semantic relation.

\paragraph{Neural retrieval and ranking.} We perform dense retrieval for the query $\vec{h^q}$, with a k-NN search over the pre-indexed sentence representations.  These results are substantially more diverse than the initial set returned by the syntactic query.

\paragraph{Argument Identification.} We encode each retrieved sentence using (Sci)BERT, and use the alignment model described in Section \ref{sec:alignment} to align spans over the retrieved sentences to the captured spans in the symbolic result set. The alignment process operates over contextualized span representations, hopefully capturing both entity type and semantic frame information.

The system returns a syntactically and lexically diverse set of results, with marked capture spans.

\section{Argument-identification via Alignment}
\label{sec:alignment}
The dense neural retrieval over the averaged query vector results in topically-related sentences. To obtain the full benefit of extractive search, we need to provide span annotations over the  sentences.  This is achieved via a span alignment model which is trained to align 
semantically corresponding spans across sentences. At query time, we apply this model to align the marked spans over the first syntactic-query result, to spans over the neurally-retrieved sentences.

The alignment model is pre-trained over a diverse set of relation, with the intent of making it a general-purpose alignment model. We describe the model architecture, training data, and training procedure.

\paragraph{The argument-alignment task.}
The user marked in the query $q$ a two spans, denoted as $ARG_1$ and $ARG_2$.  
Given a sentence (a dense retrieval result) with $n$ tokens $s=w_1,...,w_n$, we seek a consecutive sequence of tokens $w_{i:j}$ corresponding to $ARG_1$, and another consecutive sequence of tokens $w_{k:\ell}$ corresponding to $ARG_2$.

\noindent For example, consider the query:
\begin{center}
\emph{\arga{virus} \textbf{infection causes} a \argb{condition}}
\end{center}
In which the span $ARG_1$ corresponds to a kind of infection, and $ARG_2$ corresponds to the outcome of the infection.

\noindent The syntactic query may return a result such as:\\[0.1em]
\emph{The infection of \arga{SARS-CoV-2} causes \argb{fever}.}\\[0.3em]
\noindent While a neural result might be:\\[0.3em]
\indent \emph{It has been noted that headaches are one side effect of Flu infection.}\\[0.3em]
We would like to align \textit{Flu} to $ARG_1$ (SARS-COV-2) and \emph{headaches} to $ARG_2$ (fever).

\paragraph{Training and evaluation data creation.}
\label{sec:alignment-data}

To train an alignment model in a supervised setting, we need a training set that consists of pairs of sentences, both corresponding to the same relation, with arguments marked only on the first sentence. We use SPIKE for the generation of this dataset. We introduce a resource that contains 440 manually-curated SPIKE queries in the biomedical domain, divided into 67 unique relations, s.t. each relation is expressed via at least 2 syntactically-distinct queries. For instance, for the relation \textit{molecules and their chemical derivatives}, we include the following patterns, among others:
\\[0.5em]
- \arga{Something}, a \argb{Purine} \textbf{derivative}.\\
- \arga{Something}, a \textbf{derivative} of \argb{Purine}.\\ 
- \arga{Purine} \textbf{derivative such as} \argb{something}.\\[0.5em]
\noindent  We ran each SPIKE query, collect the results, and then construct a dataset that consists of randomly-sampled pair of results ($s_1^R$, $s_2^R$) for each relation $R$ of the 62 relations. This process resulted in a training set of 95,000 pairs of sentences, and a development set of 15,000 pairs of sentences, where each sentence has marked argument spans.\footnote{
We focused our efforts on maintaining high syntactic diversity alongside high topical relevance for each relation, and aimed for the patterns to cover a large set of biomedical relations.
The relations in the development set are randomly chosen subset of all relations, and are \emph{disjoint} from the relations included in the training set.}

\paragraph{Model architecture and training.}
\label{sec:alignment-model}
\begin{figure}[t]
\centering
\includegraphics[width=0.99\columnwidth]{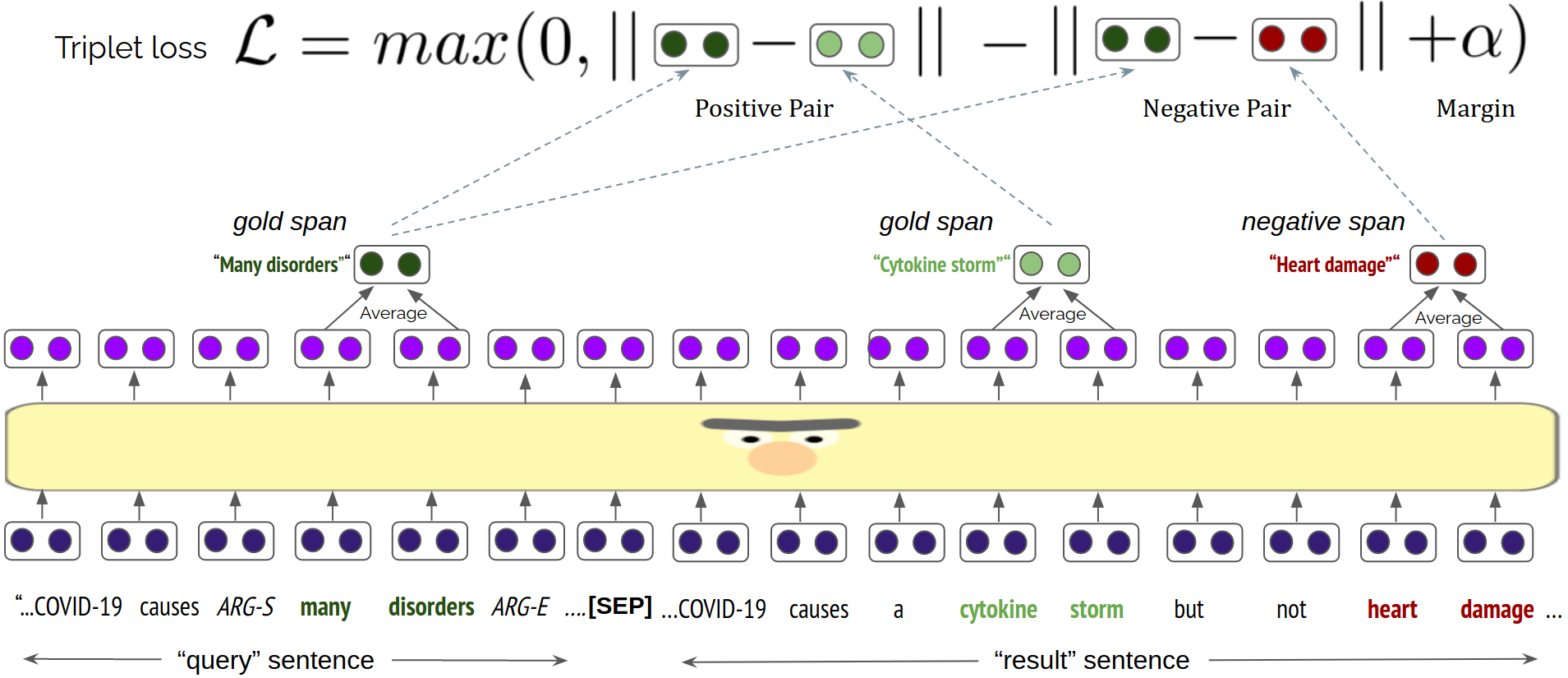}
\caption{Illustration of the argument-alignment model. We choose corresponding arguments (``many disorders" and ``cytokine storm") from the two sentences. We represent all possible spans of words, and choose the negative example to be the closest wrong span under euclidean distance (here, ``heart damage"). The triplet objective encourages the corresponding argument to be closer to each other than to the wrong span.}
\label{fig:triplet}
\end{figure}

We adopt a contrastive finetuning approach for the argument alignment task (Figure \ref{fig:triplet}). In training, the model is fed with two sentences $s_1$ and $s_2$, belonging to the same relation. On one of the sentences, we mark the argument spans using special ARG tokens. We derive contextualized representations of all consecutive spans of tokens, and contrastively train the matching spans to be more similar to each other than to any other span. 

We begin with the pretrained SciBERT model, with an additional linear layer that maps the representations to dimensionality of 64. On each training iteration we feed to the model the two sentences with arguments marked on one of them, and collect the last-hidden-layer-representations of all tokens.

    Then, we construct the representations of the two arguments in the first sentence, $\vec{h}_{arg_1}^{s_1}$ and $\vec{h}_{arg_2}^{s_1}$, by averaging the BERT representations of the tokens included in those spans. We similarly construct representations of all possible consecutive spans of tokens up to length 9 in the second sentence. The ``hardest" negative spans are identified: those are the two representations $\vec{h^{s_2, -}_{arg_1}}$ and $\vec{h^{s_2, -}_{arg_2}}$, which do not correspond to the captures in the first sentence, yet are most similar to them by euclidean distance. Those are pushed away using a triplet loss objective \cite{DBLP:conf/nips/SchultzJ03,DBLP:journals/jmlr/ChechikSSB10}:


\[
    \mathcal{L} = max(0, ||\vec{h}_{arg1}^{s_1}-\vec{h}_{arg1}^{s_2}|| - ||\vec{h}_{arg1}^{s_1}- \vec{h^{s_2, -}_{arg1}}|| + \alpha)
\]
And similarly for $arg_2$. This objective encourages the gold span in $s_1$ to be closer to the gold span $s_2$ than to any other span, with a margin of at least $\alpha$; we take $\alpha=1$ and train for 50 epochs with the Adam optimizer \cite{adam}. 

In inference time, we take $s_1$ to be an arbitrary (single) result of the syntactic query, and take $s_2$ to be any of the neural search results. For each $s_2$, we collect the spans having the least distance to the spans in $s_1$ (as provided by the SPIKE system).

\section{Evaluation}
\label{sec:alignment-evaluation}
\paragraph{Retrieval quality.}
To simulate a real-world extraction scenarios, we randomly chose 11 types of relation not included in the training set, with one randomly-selected syntactic pattern per relation. We augmented those patterns, and collected the 10 top-ranked augmented results, as well the 10 results ranked in places 90-100. We manually evaluated the relevancy of the 20 results per relation, resulting in 240 test sentences in total. In case they were relevant, we also marked the capture spans.

\paragraph{Results.} Overall, 72.2\% of the results were relevant to the relation, with 75.0\% relevant in the top-10 group and 69.5\% relevancy in the sentences ranked 90-100. In Table \ref{tab:evaluation-relevancy} with provide the results per relation. Relevancy is not uniform across relations: certain relations focusing on numerical information -- such as \textit{duration of a disease} and \textit{percentage of asympotatmic cases in a disease} had very low accuracy: the results often focused on similar but different numerical information such as \textit{``The median time to the onset of the infection was 95 days"} for \textit{duration of a disease}, and \textit{``Between 10 \% and 20 \% of the world population is infected each year by the influenza virus"} for \textit{percentage of asympotatmic cases}. In contrast, for the others relations, many of the results are relevant, and are characterized by high syntactic diversity, generalizing beyond the syntactic structure of the original symbolic query. 

\begin{table}[t!]
\centering
\begin{scalebox}{0.8}{
\begin{tabular}{lr}
\toprule
                     Relation &  \% Relevant \\
                     \hline
\midrule
             Disease-duration &      25.000 \\
         Lacunas in knowledge &     100.000 \\
      Conditions without risk &      77.273 \\
              Isolation place &     100.000 \\
      Percentage asymptomatic &       9.091 \\
                     Symptoms &      85.000 \\
          Potential treatment &      95.455 \\
 Immunutherapies and diseases &      86.364 \\
            Persistence-place &      82.609 \\
                Pretreatments &      54.545 \\
              Involved organs &      77.273 \\
\bottomrule
\hline

\end{tabular}
}
\end{scalebox}
\caption{Relevance scores (manual) by relation type. \label{tab:evaluation-relevancy}}
\end{table}

\paragraph{Alignment quality.}

To evaluate the quality of the alignment, we generate a test set from the 240 manually-annotated sentences mentioned above, by randomly sampling 1,240 pairs of sentences that belong to the same relation, and are both relevant. We keep the gold argument marking on the first sentence, omit it from the second, and have the model predict the corresponding captures. As evaluation measure, we calculate the percentage of cases where the gold argument are a subset of the predicted arguments, or vice verca.    

\paragraph{Results.} In total, 73.8\% of the arguments are aligned correctly. When considering only cases where both arguments were correctly aligned as a success, accuracy drops to 58\%. Note, however, that the captures are often multi-word expressions, and the choice of span boundaries is somewhat arbitrary, and does not take into account conjunctions or cases where possible distinct spans can be regarded as corresponding to a capture in the first sentence, and multiple valid captures that often exist within a single sentence.  

\paragraph{Comparison with symbolic extractive search.}
How do the results of the neural extractive search differ from the results of directly applying a symbolic rule based solution? To compare the systems we choose another 3 development relations, ``is a risk factor for COVID-19", ``COVID-19 spreading mechanisms" and ``potential treatment for COVID-19". For each of these relations we try out 2-3 syntactic SPIKE queries and choose the best one as a representative query. We then use the query as input for both SPIKE and for neural search \cameraReadyFootnote{We use the following SPIKE queries: \textit{arg1:something is a \$risk \$factor for arg2:[w]COVID-19}, \textit{arg1:[w]COVID-19 \$can \$spread \$by arg2:droplets} and \textit{arg1:something is a \$risk \$factor for arg2:[w]COVID-19} for the relations ``is a risk factor for COVID-19", ``COVID-19 spreading mechanisms" and ``potential treatment for COVID-19", respectively. We filter the neural-search results in remove the results not including the key word ``COVID-19".}. 

\begin{table}[]
\resizebox{\columnwidth}{!}{%
\begin{tabular}{lllll}
\hline
 & \multicolumn{2}{l}{SPIKE} & \multicolumn{2}{l}{Neural Extractive Search} \\
 & \#Caputres & \%Accuracy & \#Caputres & \%Accuracy \\ \hline
spreads by & 5 & 83\% & 40 & 96\% \\
potential treatment & 14 & 80\%  & 55 & 67.6\% \\
risk factor & 57 & 89\% & 44 & 83\% \\ \hline
\end{tabular}%
}
\caption{Comparing result count and accuracy between symbolic and neural extractive search \label{tab:comparison-to-spike}}
\end{table}

As shown in Table \ref{tab:comparison-to-spike}, for two of the three relations, \emph{spread} by and \emph{potential treatment}, neural search has been effective in significantly improving recall while maintaining relatively high precision. For the third relation, \emph{risk-factor}, neural search did not show benefit but did not lag far behind. We hypothesize that this is due to this relation appearing many times in the data and in less diverse ways compared to the other relations, allowing a symbolic pattern to accurately extract it. Importantly, these data suggest that the neural search system is less sensitive to the exact relation and query used and that in some cases it also significantly improves performance.


\section{Example Search}
\label{sec:example}
We demonstrate the system via an example where one aims to find sentences containing information on compounds and their origin (e.g. plant-derived, isolated from soil, etc.). We start with the query:
\begin{center}
\emph{\argas{Something} is a \textbf{drug} \textbf{extracted} from \argbs{plants}}. 
\end{center}
The syntactic yields only few results, all of them are relevant. Among the results: \\
-\emph{\underline{Colchicine} is a drug extracted from \underline{Colchicum} \underline{autumnale}.}\\[0.3em]
-\emph{\underline{Berbamine} is an experimental drug extracted from \underline{a shrub} native to Japan, Korea, and parts of China}\\[0.3em]
-\emph{\underline{Taxol}, isolated from \underline{Taxomyces andreanae} , is the most effective and successful anticancer drug extracted from endophytic fungi to date .}
Figure \ref{fig:screenshot} shows the output (top results) of the neural system.
The neural results are notably more diverse. 
While the syntactic results follow the pattern ``X extracted from Y", the neural results are both lexically and syntactically diverse: the explicit descriptor ``a drug" is absent at times; the verbal phrase ``extracted from [a plant]" is sometimes replaced with the paraphrases ``found in [a plant]" and ``[is a] botanical extract"; and the third result contains an unreduced relative clause structure.

Several additional results are presented below:

\noindent - \emph{\underline{Allicin} is the major biologically active component of \underline{garlic}.}

\noindent - \emph{Berberine is an isoquinoline \underline{alkaloid} that has been isolated from Berberis \underline{aquifolium}.}

\noindent - \emph{\underline{Phillyrin} ( Phil ) , the main pharmacological component of Forsythia \underline{suspensa}, possesses a wide range of pharmacological activities .}

\noindent - \emph{Dimethyl cardamonin ( DMC ) is the active \underline{compound} isolated from the leaves of \underline{Syzygium} samarangense.}

\noindent - \emph{\underline{Triostin} is a well-known natural product with antibiotic , \underline{antiviral}, and antitumor activities .}

Note that the last two examples demonstrate \emph{failure modes}: in the the fourth example, the model failed to identify \textit{Dimethyl cardamonin} as an argument; and in the last sentence there is no clear capture corresponding to the second argument.

Finally, we perform an aggregation over the predicted captures (Fig \ref{fig:screenshot}, right-pane), allowing the user to quickly get a high-level overview of the interactions. From our experience, users are mostly interested in this table, and turn to the text as support for validating interesting findings.

\section{Limitations of the neural approach} While we find the neural approach to be very effective, we would also like to discuss some of its limitations. First, speed and scalability are still lagging behind that of symbolic search systems: dense retrieval systems do not yet scale as well as symbolic ones, and running the (Sci)BERT-base argument-aligner for each candidate sentence is significantly slower than performing the corresponding similarity search. While the user can see the first results almost immediately, getting extractions from thousands of sentences may take several minutes. We hope to improve this speed in future work.

In terms of accuracy, we find the system to be hit-or-miss. For many symbolic queries we get fantastic resutls, while for others we observe failures of the dense retrieval model, or frequent failures of the alignment model, or both.  For effective incorporation in a user-facing system, we should---beyond improvements in retrieval and alignment accuracy---be able to predict which queries are likely to yield poor results, and not extend them with fuzzy neural matches.


\section{Conclusions}
\label{sec:conclusions}
We presented a system for neural extractive search. While we found our system to be useful for scientific search, it also has clear limitations and areas for improvement, both in terms of accuracy (only 72.2\% of the returned results are relevant, both the alignment and similarity models generalize well to some relations but not to others), and in terms of scale. We see this paper as a beginning rather than an end: we hope that this demonstration will inspire others to consider the usefulness of the neural extractive search paradigm, and develop it further.

\section*{Acknowledgements}
This project received funding from the Europoean Research Council (ERC) under the Europoean Union's Horizon 2020 research and innovation programme, grant agreement No. 802774 (iEXTRACT).

\bibliography{acl2020}
\bibliographystyle{acl_natbib}
 
\end{document}